\documentclass[journal,comsoc]{IEEEtran}

\usepackage[T1]{fontenc}
\usepackage[pdftex]{graphicx}
\usepackage{float}
\usepackage{subfigure}
\usepackage{color}
\usepackage{subfigure}
\usepackage{multirow}
\usepackage{hyperref}

\ifCLASSINFOpdf

\else

\fi

\usepackage{amsmath}

\usepackage[cmintegrals]{newtxmath}

\begin{document}

\title{A Survey on Foundation-Model-Based Industrial Defect Detection}

\author{Tianle Yang$^{*}$, Luyao Chang$^{*}$, Jiadong Yan, Juntao Li, Zhi Wang, Ke Zhang

\thanks{Corresponding author: K. Zhang is with  Soochow University, Suzhou, China. (e-mail:kzhang19@suda.edu.cn).}
\thanks{T. Yang, J. Yan, and J. Li are with Soochow University, Suzhou, China. (e-mail:tlyang@stu.suda.edu.cn, jdyan24@stu.suda.edu.cn, ljt@suda.edu.cn).}
\thanks{Z. Wang is with Shenzhen International Graduate School, Tsinghua University, Beijing, China (e-mail: wangzhi@sz.tsinghua.edu.cn).}
\thanks{L. Chang is with Wuhan University of Science and Technology, Wuhan, China. (e-mail:changluyao001@163.com).}
\thanks{\textit{(T. Yang  and L. Chang contributed equally to this paper.)}}
}

\maketitle

\begin{abstract}
As industrial products become abundant and sophisticated, visual industrial defect detection receives much attention, including two-dimensional and three-dimensional visual feature modeling. Traditional methods use statistical analysis, abnormal data synthesis modeling, and generation-based models to separate product defect features and complete defect detection. Recently, the emergence of foundation models has brought visual and textual semantic prior knowledge. Many methods are based on foundation models (FM) to improve the accuracy of detection, but at the same time, increase model complexity and slow down inference speed. Some FM-based methods have begun to explore lightweight modeling ways, which have gradually attracted attention and deserve to be systematically analyzed. In this paper, we conduct a systematic survey with comparisons and discussions of foundation model methods from different aspects and briefly review non-foundation model (NFM) methods recently published. Furthermore, we discuss the differences between FM and NFM methods from training objectives, model structure and scale, model performance, and potential directions for future exploration. Through comparison, we find FM methods are more suitable for few-shot and zero-shot learning, which are more in line with actual industrial application scenarios and worthy of in-depth research.
\end{abstract}


\IEEEpeerreviewmaketitle

\section{Introduction}

\IEEEPARstart{V}{isual} defect detection \cite{pang2021deep}, which is also called visual anomaly detection, is a key application area of artificial intelligence algorithms. This task plays a crucial role in ensuring the quality of industrial products. Traditional industrial anomaly detection algorithms \cite{bergmann2018improving,gong2019memorizing,liu2021unsupervised,deng2022anomaly} focus on modeling the statistical distribution of normal features and detecting anomalies by analyzing the deviations in input samples from these learned patterns. To enhance the model’s ability to identify anomalous patterns, some methods \cite{liu2023simplenet,liang2024tocoad} further explore contrastive learning mechanisms \cite{hu2024comprehensive} between normal and abnormal features. These methods typically rely on a large amount of high-quality training data to establish reliable feature distributions and contrastive relationships. However, in real industrial scenarios, it is challenging to acquire specific high-quality training data due to the diversity and complexity of products and defects \cite{bergmann2019mvtec,zou2022spot,wang2024real}. For example, in chip defect detection, there are many types of chips and numerous defect categories, including structural defects and texture defects, making it difficult to collect data for various products and defects. In such cases, traditional models struggle to achieve satisfactory detection results. Recently, with the release of foundation models in vision and language, such as CLIP \cite{radford2021clip}, GPT \cite{zhu2023minigpt4,yang2023gpt4v} and SAM \cite{kirillov2023sam}, industrial defect detection algorithms based on these models have made significant progress in both 2D and 3D visual environments \cite{rani2024advancements}, particularly in few-shot and zero-shot scenarios where data are limited. This has received a great deal of attention. \textbf{The foundation models themselves possess strong capabilities in understanding general vision and language, making it an important issue to explore how to effectively apply their foundational knowledge to industrial detection problems without additional training samples and annotations. }
We categorize the application of different foundation models in 2D and 3D industrial defect detection as follows:

\begin{figure*}[ht]
    \centering
    \includegraphics[width=1\linewidth]{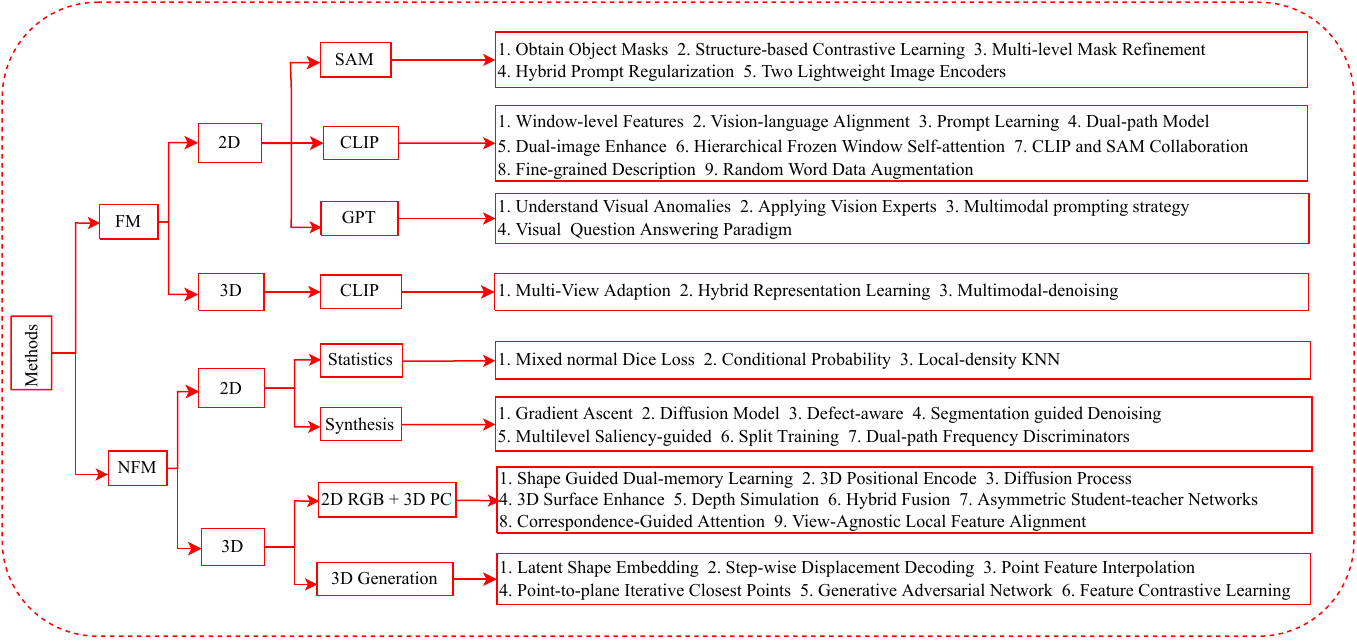}
    \caption{Organization of surveyed methods. We categorize the methods under investigation into two main categories: foundation models and non-foundation models. Each category is further divided into 2D and 3D scenarios. The foundation-model-based methods primarily include methods based on SAM, CLIP, and GPT, while non-foundation-model-based methods are classified into static methods, synthesis-based methods, methods combining 2D RGB and 3D point clouds, and 3D generative methods. Finally, we present the latest methods collected in this survey.}
\end{figure*}

\begin{enumerate}
    \item SAM-2D: \textbf{Application of visual prior knowledge}. As a powerful foundational model for visual segmentation, SAM provides semantic prior information acquired through extensive pre-training on vast amounts of data, significantly enhancing the accuracy of industrial defect detection. In 2D industrial defect detection tasks based on SAM, researchers have developed various methods \cite{li2024sam,cao2023segment,peng2024sam,liu2024unsupervised,li2024clipsam,yang2024spt} to prompt SAM specifically for industrial scenarios. Additionally, object matching based on the masks generated by SAM is used to identify defect regions.
    \item CLIP-2D: \textbf{Semantic matching of short texts and images}. Image-text foundation models such as CLIP demonstrate fine-grained image-text matching. This ability effectively links subtle visual cues with descriptive text, so it is especially beneficial for defect detection. In 2D industrial defect detection tasks based on CLIP \cite{jeong2023winclip,zhou2023anomalyclip,cao2025adaclip,qu2024vcp,deng2024simclip,chen2024clipad,zuo2024clipfsac,li2024clipsam,hu2024sowa,chen2023april,li2024promptad,gu2024filo,zhang2024dual}, it is essential to design and learn suitable text prompts while aligning image information at a fine-grained level to further enhance performance. The design of text prompt templates has been extensively studied.
    \item GPT-2D: \textbf{Long text semantic prior}. Large language models like GPT can generate long-form descriptions, making them very suitable for complex scenarios that require detailed explanations and structured descriptions. Therefore, a key challenge in GPT-based 2D industrial defect detection methods \cite{gu2024anomalygpt,cao2023towards,xu2024customizing,li2023myriad,zhu2024alfa,zhang2024gpt,zhang2024logicode} is designing prompts to obtain comprehensive text descriptions and effectively leveraging the textual information.
    \item CLIP-3D: \textbf{Short-text image semantic matching prior applied to cross-dimensional vision tasks}. 3D defect detection faces greater challenges due to its complex spatial information. To address this, image-text foundation models like CLIP offer a promising solution through cross-modal information complementarity \cite{zuo2024clip3d,zhou2024pointad,wang2024m3dm}. These models effectively combine visual information with textual descriptions, thereby enabling more precise high-dimensional spatial modeling that captures complex defects in 3D structures.
\end{enumerate}

Although FMs demonstrate promising application prospects in industrial defect detection, NFM methods still possess irreplaceable advantages in specific application scenarios due to their smaller parameter sizes and higher computational efficiency. Based on this, this paper also provides a review of NFM methods, including 2D statistical modeling \cite{zhang2024sofs,bae2023pni,lyu2024reb,yao2023bgad,qian2024friend}, 2D anomaly data synthesis \cite{chen2025glass,li2024adabldm,zhang2024realnet,jiang2024cagen,hu2024anomalyxfusion,hu2024anomalydiffusion,duan2023dfmgan,zhang2023destseg,qin2024cutswap,lin2024split,bai2024dfd,chen2024pbas}, 2D/3D cross-modal knowledge distillation \cite{chu2023shape,cao2024cpmf,horwitz2023back,fuvcka2025transfusion,zavrtanik20243dsr,wang2024m3dm,rudolph2023ast}, and algorithms based on 3D generative models \cite{zhou2025r3d,zhao2024pointcore,zuo2024clip3d,liu2024uni,zhu2024group3ad}. \textbf{We believe that these methods can provide effective insight for FM methods and some of them can be applied to FM models}. In addition, we systematically compare the differences between foundation and non-foundation approaches in terms of application scenarios, algorithm framework focus, detection performance, model complexity, and future development directions. Key areas for potential breakthroughs in both approaches are also highlighted. This paper aims to provide researchers and engineers with information on selecting the appropriate research methods for different scenarios and to offer valuable perspectives on the future development of industrial defect detection.

The organization of this survey paper is as follows. First, in Section 1, we introduce the challenges posed by FM in industrial defect detection, followed by a discussion of the mainstream methods currently adopted. In Section 2, we provide a detailed comparison between FM and NFM methods, focusing on their differences in training objectives, model architectures, algorithm framework and performance. Then we give an overview of the different types of FM methods applied to both 2D and 3D industrial defect detection in Section 3. Section 4 discusses the key approaches of NFM methods and insights they provide for FM methods. Finally, in Section 5, we examine the ongoing challenges faced by large models and highlight potential future directions for further exploration. A detailed organization of the methods we investigate is also shown in Figure 1.

\section{Comparison of FM and NFM methods}

With the diversification of industrial detection demands, the differences in model training objectives, structures, scales, and performance have become key factors influencing the choice of methods. The following comparison analyzes the performance of FM and NFM in industrial anomaly detection from the perspectives of training objectives, model structure and scale, algorithm framework and performance. A summary of the comparison is shown in Figure 2.

\begin{figure*}[t]
    \centering
    \includegraphics[width=1\linewidth]{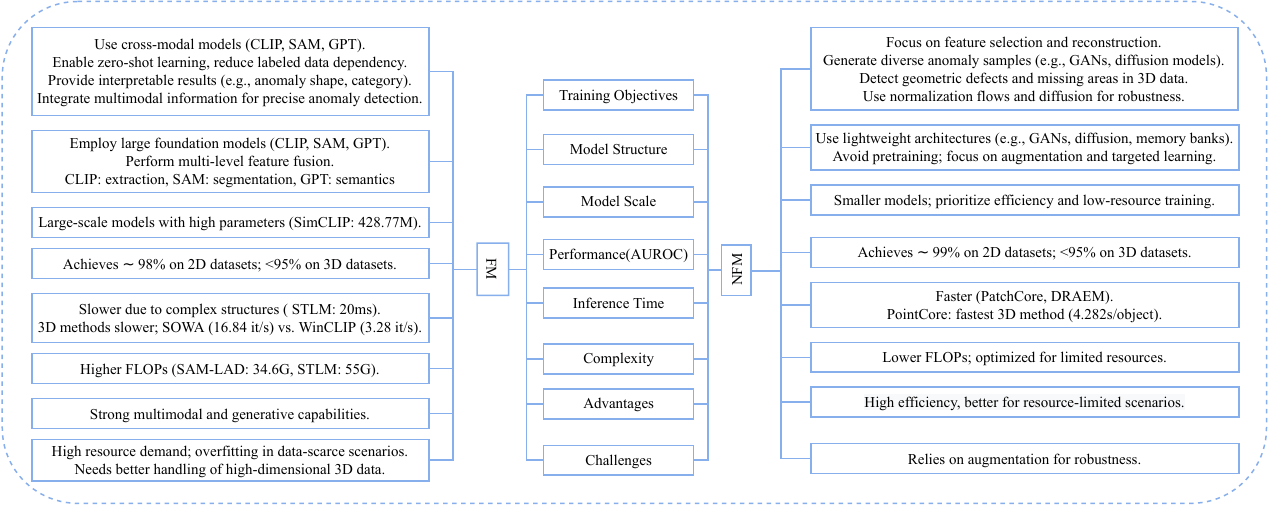}
    \caption{A summary of the comparison between FM and NFM methods. We conduct a systematic comparison of the FM and NFM methods from the following 5 aspects: 1) Model Training Objectives. 2) Model Structure. 3) Model Scale. 4) Model Performance (AUROC Performance, Inference Time, and Computational Complexity). 5) Advantages and Challenges.}
\end{figure*}

\subsection{Model Training Objectives}
FM and NFM exhibit significant differences in data requirements, training methods, computational resources, and the breadth of feature learning, which consequently leads to differences in their training objectives. 2D FM methods (e.g., CLIP, SAM, GPT) leverage the cross-modal capabilities\cite{liu2024speech,tu2025self,li2024learning} of vision-language models to improve the accuracy and efficiency of anomaly detection. Their main objectives include:
1) Identifying unknown anomaly categories through unsupervised or zero-shot learning, reducing the dependence on labeled data; 2) Generating interpretable detection results that describe anomalies in terms of color, shape, and category; 3) Enhancing accuracy by integrating specific anomaly observation modules with the FM, addressing complex anomalies; 4) Improving model scalability and adaptability, enabling rapid adaptation to different industrial scenarios; 5) Integrating multimodal information to achieve precise anomaly localization and identification. 
3D FM methods focus on the geometric features of point cloud data \cite{arav2024evaluating,ye2024po3ad} and multi-view fusion \cite{hao2024network}, addressing issues of incomplete data and noise interference \cite{dai2017scannet,uy2019revisiting}. They perform classification and segmentation through multi-view rendering, while also handling inconsistencies between multimodal data. 

In contrast, 2D NFM methods rely on traditional network architectures, utilizing techniques such as GANs \cite{al2024enhanced} and diffusion models \cite{bhosale2024anomaly} to generate diverse anomaly samples to compensate for insufficient data. They emphasize feature selection and reconstruction 
 \cite{kim2024rethinking,yao2025glad,rafiee2024dcor,patra2024revisiting}strategies. 3D NFM methods focus on detecting geometric defects and missing areas in point cloud data, using efficient architectures to reduce computational overhead. They also employ innovative techniques, such as normalization flows \cite{lee2024gdflow,zhou2024vq} and diffusion-based reconstruction mechanisms, to enhance accuracy and robustness, avoiding dependence on design files or model libraries.

In summary, FM methods focus on \textbf{cross-modal learning and generative capabilities}, excelling in data-scarce scenarios and suitable for multi-task and multi-domain detection. In contrast, NFM methods emphasize \textbf{feature selection, computational efficiency, and data synthesis}, making them more suitable for resource-constrained environments.

\begin{figure*}[t]
    \centering
    \includegraphics[width=1\linewidth]{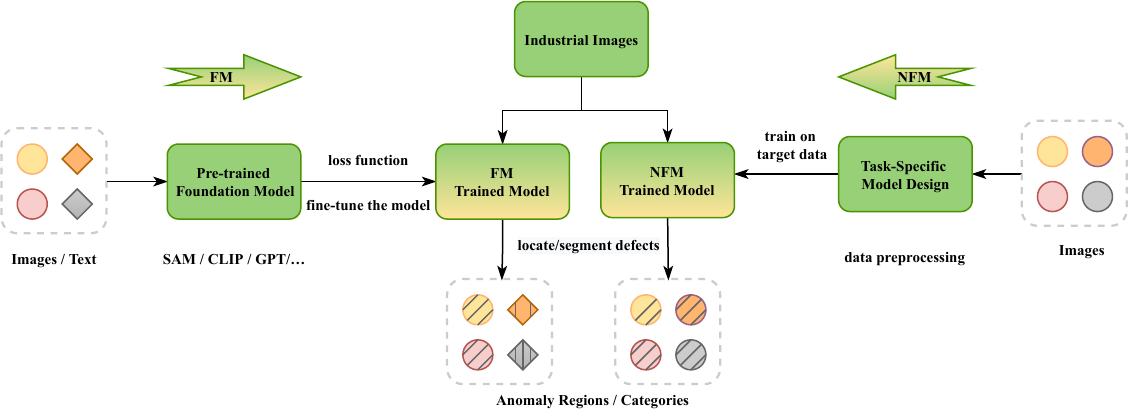}
    \caption{The left branch is framework of FM methods and the right one is of NFM methods. FM methods are primarily based on FM such as SAM, CLIP and GPT. During training, FM methods design appropriate loss functions to fine-tune the pre-trained foundational models, adapting them to the industrial defect detection domain. In contrast, NFM methods focus on designing task-specific models based on lightweight or specialized network architectures. Some NFM methods also design anomaly synthesis strategies to supplement training data.}
\end{figure*}

\subsection{Model Structure and Scale}
\subsubsection{Model structure}
FM methods rely on powerful \textbf{vision-language collaborative mechanisms}, integrating large-scale foundational models such as CLIP, SAM, and GPT. These models employ multi-level feature fusion to establish a collaborative workflow: CLIP performs multi-modal feature extraction and alignment on image and point cloud data, SAM carries out fine-grained segmentation to isolate potential anomaly regions, and GPT provides semantic understanding and description of the detection results, assisting users in quickly obtaining analytical conclusions. To address the challenges of few-shot and zero-shot learning \cite{chen2024survey}, CLIP’s pre-trained knowledge enables effective inference on unlabeled data, thereby enhancing the generalization ability of the detection model. NFM methods mainly include Teacher-Student Architecture \cite{sun2024memoryless,deng2024structural,chen2024filter,liu2024unlocking}, Distribution Map, Memory Bank \cite{xing2023visual}, Autoencoder-based \cite{liang2024automated}, GAN-based, Transformer-based, and Diffusion-based frameworks. These approaches do not rely on large-scale data or pretraining tasks, focusing more on \textbf{local feature selection, sample generation, and augmentation}. Their aim is to optimize feature learning and anomaly detection capabilities with limited data.
\subsubsection{Model scale}
FM methods typically rely on \textbf{large parameter sizes}, utilizing complex network architectures and cross-modal learning to handle intricate anomaly detection tasks. This results in higher training times and computational resource demands. For example, SimCLIP \cite{deng2024simclip} has parameter sizes of 428.77M. In contrast, NFM methods have smaller parameter sizes and primarily optimize models through efficient \textbf{feature selection, adversarial training, and self-supervised learning}. These methods can achieve more efficient training in resource-constrained environments. Since fast inference is an inevitable trend, the newly published FM methods are trying to explore ways to accelerate inference. For example, SAM-based STLM \cite{li2024sam} requires only 16.56M for inference, making it one of the most efficient methods.

\subsection{Framework}
The frameworks of FM methods and NFM methods are shown in Figure 3. FM methods primarily leverage \textbf{the prior knowledge embedded in foundation models}, which have been pre-trained on large-scale general-purpose datasets and possess strong feature representation capabilities. Consequently, fine-tuning these models often requires only a small number of samples. Different types of foundation models, such as SAM and CLIP, can process data of different modalities, including images and textual information. During training, FM methods focus on designing suitable loss functions to adapt foundation models more effectively to anomaly detection tasks in industrial applications. Ultimately, the fine-tuned models achieve accurate segmentation or localization of anomalous regions in industrial images.

NFM methods focus on \textbf{designing task-specific models}. For example, reconstruction-based anomaly detection methods train a model that can accurately reconstruct normal data by learning the reconstruction process. During data preprocessing, some methods use anomaly synthesis strategies to expand the dataset since anomaly samples are rare. By training the model with the target data, it gradually improves and ultimately generates a model specialized in detecting or segmenting anomalous regions.

\subsection{Model Performance}
\subsubsection{AUROC performance}
As shown in Table 1, using the commonly employed MVTec dataset as an example, 2D NFM methods generally achieve AUROC values close to 99\%, performing the best. In contrast, some 2D FM methods have AUROC values around 98\%. In 3D methods, both FM and NFM exhibit average AUROC values below 95\%. \textbf{This indicates that 2D methods outperform 3D methods overall, and NFM is currently more mature than FM in this context.}

\begin{table*}
\centering
\caption{A brief summary and overview of different FM and NFM methods. The numbers of performance are all copied from their original paper. Specifically, “Performance” denotes the AUROC metric on the dataset shown behind.}
\resizebox{\textwidth}{!}{%
\begin{tabular}{c|c|c|c|c|c|c} 
\hline
Category & Sub-category & Method & Description & Publication & \multicolumn{2}{c}{Performance}\\ 
\hline
\multirow{29}{*}{Foundation}& \multirow{6}{*}{2D SAM Based}& ClipSAM \cite{li2024clipsam}& Hierarchical mask refinement with multi-level prompts&             ARXIV 2024&               92.3 &\multirow{28}{*}{MVTec AD}\\
 \multirow{30}{*}{Model}& & UCAD \cite{liu2024unsupervised}
& Structure-based contrastive learning with SAM& AAAI 2024& 93.0 &\\
 \multirow{31}{*}{Method}& & SAM-LAD \cite{peng2024sam}
& Use SAM to obtain object masks of the query and reference images and extract object features for matching& ARXIV 2024&98.4 &\\
 & & SAA+ \cite{cao2023segment}
& Hybrid prompt regularization& ARXIV 2023&- &\\ 
 & & STLM \cite{li2024sam}
& Utilize SAM as a teacher to guide student networks& ARXIV 2024&98.26 &\\
& & SPT \cite{yang2024spt}& Adapt SAM to better understand the relationships between different regions in the image& AAAI 2025& -&\\
\cline{2-6}& \multirow{15}{*}{2D CLIP Based}&WinCLIP \cite{jeong2023winclip}&Compositional prompt ensemble, reference association method&CVPR 2023&93.1 &\\ 
& &AnoCLIP \cite{deng2023anovl}&Local-aware visual tokens, domain-aware prompting, test-time adaptation method&ARXIV 2024&-
 &\\ 
& &AnomalyCLIP \cite{zhou2023anomalyclip}&An object-agnostic text prompt template, global abnormality loss function&ICLR 2024&-
 &\\ 
& &AdaCLIP \cite{cao2025adaclip}&Hybrid (static and dynamic) learnable prompts, hybrid-semantic fusion module&ECCV 2024&-
 &\\ 
& &VCP-CLIP \cite{qu2024vcp}&Visual context prompting model&ARXIV 2024&-
 &\\ 
& &SimCLIP \cite{deng2024simclip}&Multi-hierarchy vision adapter, implicit prompt learning, prior-aware optimization algorithm&ARXIV 2024&95.3
 &\\ 
& &CLIP-AD \cite{chen2024clipad}&Distribution of the text prompts, facilitate alignment via a linear layer&ARXIV 2023&-
 &\\ 
& &CLIP-FSAC \cite{zuo2024clipfsac}&Two-stage training strategy, visual-driven text features, fusion-text matching task&IJCAI 2024&95.5
 &\\ 
& &ClipSAM \cite{li2024clipsam}&CLIP and SAM Collaboration, unified multi- scale cross-modal interaction, multi-level mask refinement&ARXIV 2024&-
 &\\ 
& &SOWA \cite{hu2024sowa}&Hierarchical frozen window self-attention, dual Learnable Prompts&ARXIV 2024&-
 &\\ 
& &SAA+ \cite{cao2023segment}&Hybrid prompts, domain expert knowledge and target image context&ARXIV 2023&--
 &\\ 
& &APRIL-GAN \cite{chen2023april}& Employ a combination of state and template ensembles, memory bank-based approach &ARXIV 2023&92.0
 &\\ 
& &PromptAD \cite{li2024promptad}&Prompt learing, semantic concatenation, explicit anomaly margin&CVPR 2024&94.6
 &\\ 
& &FiLo \cite{gu2024filo}&Fine-grained description, learnable vectors, position-enhanced high-quality localization method&ARXIV 2024&-
 &\\ 
& &Dual-Image Enhanced CLIP \cite{zhang2024dual}&Dual image feature enhancement, test-time adaption with pseudo anomaly synthesis&ARXIV 2024&-
 &\\ 
\cline{2-6}& \multirow{7}{*}{2D GPT Based}&AnomalyGPT \cite{gu2024anomalygpt}&Lightweight and visual-textual feature-matching-based decoder, prompt embeddings&AAAI 2024&94.1
 &\\ 
& &Myriad \cite{li2023myriad}&Apply vision experts, vision expert tokenizer&ARXIV 2023&94.1
 &\\ 
& &ALFA \cite{zhu2024alfa}&Run-time prompt adaptation strategy, fine-grained aligner&ARXIV 2024&94.5 &\\ 
& &GPT-4V-AD \cite{zhang2024gpt}&Visual Question Answering paradigm, granular region division, prompt designing, Text2Segmentation method&ARXIV 2023&- &\\
& &Customizable-VLM \cite{xu2024customizing}&Enhance foundation models by integrating expert knowledge as external memory via prompting & ARXIV 2024& 82.9&\\
& &LogiCode \cite{zhang2024logicode}&Use LLMs to extract image logic and generate code for logical anomaly detection& ARXIV 2024& -&\\
\cline{2-7}& \multirow{3}{*}{3D CLIP Based}&CLIP3D-AD \cite{zuo2024clip3d}&Address both few-shot anomaly classification and segmentation without memory banks and plenty of training samples&ARXIV 2024&- &\multirow{3}{*}{MVTec3D-AD}\\ 
 & & PointAD \cite{zhou2024pointad}
 & Hybrid representation learning framework& ARXIV 2024&97.2 &\\
& & M3DM-NR \cite{wang2024m3dm}
& Use the suspected anomaly maps to achieve denoising& ARXIV 2024&94.5 &\\
\hline
\multirow{27}{*}{Non-Foundation}&\multirow{5}{*}{2D Statistic}&SOFS \cite{zhang2024sofs}
&Introduce an abnormal prior map and mixed normal Dice loss&CVPR 2024&93.3
 &\multirow{17}{*}{MVTec AD}\\ 
\multirow{28}{*}{Model}& &PNI \cite{bae2023pni}&Utilize position and neighborhood information&ARXIV 2023&99.56
 &\\ 
\multirow{29}{*}{Method}& &REB \cite{lyu2024reb}&Reduce domain and local density biases&ARXIV 2024&99.5&\\ 
& &BGAD \cite{yao2023bgad}&Strengthen the decision boundary by pulling together normal samples while pushing away anomalous samples & CVPR 2023& 99.3&\\
& &COAD \cite{qian2024friend}&Enhance model sensitivity to anomalies through controlled overfitting& ARXIV 2024& 99.9&\\
 \cline{2-6}& \multirow{12}{*}{2D Synthesis}& GLASS \cite{chen2025glass}
&                                              Anomaly synthesis based on Gaussian noise and gradient rise&             ARXIV 2024&               99.9 &\\
 & & AdaBLDM \cite{li2024adabldm}
& Latent diffusion model with feature editing& ARXIV 2024&- &\\
 & & RealNet \cite{zhang2024realnet}
& Strength-controllable diffusion anomaly synthesis& CVPR 2024&99.6 &\\
 & & CAGEN \cite{jiang2024cagen}
& Text-guided controllable anomaly generation& ICASSP 2024&97.7 &\\
 & & AnomalyXFusion \cite{hu2024anomalyxfusion}
& Multi-modal anomaly synthesis for enhanced sample fidelity& ARXIV 2024&99.2 &\\
 & & AnomalyDiffusion \cite{hu2024anomalydiffusion}
& Spatial anomaly embedding, adaptive attention re-weighting mechanism& AAAI 2024&99.2 &\\
 & & DFMGAN \cite{duan2023dfmgan}
& Use defect-aware residual blocks in StyleGAN2& AAAI 2023&- &\\
 & & DeSTSeg \cite{zhang2023destseg}
& Denoising student encoder-decoder, adaptive multi-level feature fusion& CVPR 2023&98.6 &\\
 & & CutSwap \cite{qin2024cutswap}
& Leverages saliency guidance to incorporate semantic cues& ARXIV 2023&98.0 &\\
 & & Split Training \cite{lin2024split}
& A split training strategy that alleviates the overftting issue& ARXIV 2024&98.3 &\\
 & & DFD \cite{bai2024dfd}
 & Frequency-domain analysis with dual-path frequency discriminators& ARXIV 2024&93.3 &\\
 & &PBAS \cite{chen2024pbas}&Use the compact distribution of normal sample features to guide the direction of feature-level anomaly synthesis  & TCSVT 2024& 99.8&\\
 \cline{2-7}& \multirow{7}{*}{2D RGB+3D PC}& Shape-Guided \cite{chu2023shape}
& Synergistic expert models for anomaly localization in color and shape& WACV 2024
& 94.7 
 &\multirow{7}{*}{MVTec3D-AD}\\ 
 & & CPMF \cite{cao2024cpmf}
& Combine handcrafted PCD descriptions with 
pre-trained 2D neural networks& Pattern Recognition 2023
&92.93 
 &\\
 & & Back to the Feature \cite{horwitz2023back}
& Handcrafted 3D representations with PatchCore& CVPR 2021
&97.8 
 &\\
 & & TransFusion \cite{fuvcka2025transfusion}
& Address the overgeneralization and loss-of-detail problems utilizing transparency-based diffusion& ECCV 2024&98.2 
 &\\
 & & 3DSR \cite{zavrtanik20243dsr}
& Depth-aware discrete autoencoder and the
simulated depth generation process& WACV 2024&97.8 
 &\\
 & & M3DM \cite{wang2024m3dm}
& A hybrid fusion scheme to reduce the disturbance between multimodal features and encourage feature interaction& CVPR 2023
&94.5 
 &\\
 & & AST \cite{rudolph2023ast}
& Introduce a network which compensates for wrongly estimated
likelihoods by a normalizing flow& WACV 2023
&93.7 
 &\\
 \cline{2-7}& \multirow{5}{*}{3D Generation}& R3D-AD \cite{zhou2025r3d}
& Overcome the inefficiencies due
to the memory bank module and low performance caused by incorrect rebuilds
with MAE& ECCV 2024
&73.4 
 &Real 3D-AD
\\
 & & Reg 3D-AD \cite{liu2024real3d}
& A dual-feature representation approach to preserve the training prototypes’
local and global features& NeurIPS 2023
&70.4 
 &Real 3D-AD
\\
 & & PointCore \cite{zhao2024pointcore}
& Reduce the computational cost and mismatching disturbance
in inference& ARXIV 2024
&82.9 
 &Real 3D-AD
\\
 & & Uni-3DAD \cite{liu2024uni}
& Notable adaptability to model-free industrial
products& ARXIV 2024
&-
 &MVTec 3D-AD\\
 & & Group3AD \cite{zhu2024group3ad}
 & Enhance the
resolution and accuracy of 3D anomaly detection through group level feature contrastive learning& ACM MM 2024&75.1 &Real 3D-AD\\ 
\hline
\end{tabular}
}
\label{tab:booktabs}
\end{table*}

\subsubsection{Inference time}
\textbf{Due to their large parameter sizes and complex model structures, FM methods generally require more inference time.} Although STLM \cite{li2024sam} has significantly optimized inference efficiency with an average inference time of 20ms, it still lags behind NFM methods like DRAEM \cite{zavrtanik2021draem}, FastFlow \cite{yu2021fastflow}, and PatchCore \cite{roth2022towards}, which have shorter inference times. 3D methods typically require longer inference times; however, some methods, such as SOWA \cite{hu2024sowa}, still demonstrate excellent inference speed, with a rate of 16.84 it/s (compared to 3.28 it/s for WinCLIP \cite{jeong2023winclip} and 1.82 it/s for April-GAN \cite{chen2023april}). Compared to BTF \cite{horwitz2023back}, M3DM \cite{wang2023multimodal}, and Reg 3D-AD \cite{liu2024real3d}, PointCore \cite{zhao2024pointcore} achieves the highest AUROC and is the fastest, with a mean inference time per object on Real3D-AD \cite{liu2024real3d} of 4.282s, excluding BTF \cite{horwitz2023back}(2.19s). The shape-guided \cite{chu2023shape} method has an inference time of 2.05s per sample, outperforming BTF \cite{horwitz2023back}.

\subsubsection{Computational complexity}
\textbf{FM methods typically have higher FLOPs than those of NFM methods due to their large model sizes.} For instance, STLM \cite{li2024sam} has a FLOPs of 55G. SAM-LAD  \cite{peng2024sam}, which employs transformers and upsampled feature maps, has a FLOPs of 54.7G, higher than that of CNN-based NFM methods such as AE \cite{bergmann2018improving}(5.0G) and f-AnoGan \cite{schlegl2019f}(7.7G). Addressing the computational complexity of FM has become a popular research direction. SimCLIP \cite{deng2024simclip}, optimized for inference efficiency, requires fewer FLOPs (513.75G) than SOTA prompt learning methods like CoOp \cite{zhou2022learning}(520.46G) and Co-CoOp \cite{zhou2022conditional}(520.46G) while maintaining the same parameter count. However, SimCLIP’s \cite{deng2024simclip} FLOPs are an order of magnitude higher than STLM \cite{li2024sam} and SAM-LAD \cite{peng2024sam}, as STLM \cite{li2024sam} uses distillation from a fixed SAM teacher, and SAM-LAD \cite{peng2024sam} employs FeatUp’s \cite{fu2024featup} Upsampling Factors. And both STLM \cite{li2024sam} and SAM-LAD \cite{peng2024sam} do not use foundation models during inference.

Based on the above analysis, FM methods demonstrate strong potential in complex industrial detection and cross-domain applications, thanks to their powerful multi-modal capabilities and large parameter sizes. However, challenges such as overfitting in data-scarce scenarios and inference efficiency remain bottlenecks, particularly when handling 3D high-dimensional data, which still offers ample room for exploration. NFM methods, on the other hand, rely on targeted feature extraction and efficient computation, making them more advantageous in real-time inference and industrial scenarios with limited computational resources.

\section{Foundation Model Methods}

In recent years, visual-language models have shown significant advantages in anomaly detection. These models are able to better understand and describe complex features in images by effectively combining visual information with linguistic cues. Compared to traditional anomaly detection methods, visual-language models are able to exploit rich contextual information and reduce the dependence on manual annotation and domain knowledge, thus achieving more accurate detection. In this part, three main classes of methods based on visual- language models are introduced: methods based on SAM, CLIP and GPT. In Table 1, we give a summary and overview of different
methods based on FM. And in Figure 4, the most important and popular works along the FM development are shown in the timeline.

\begin{figure*}[t]
    \centering
    \includegraphics[width=1\linewidth]{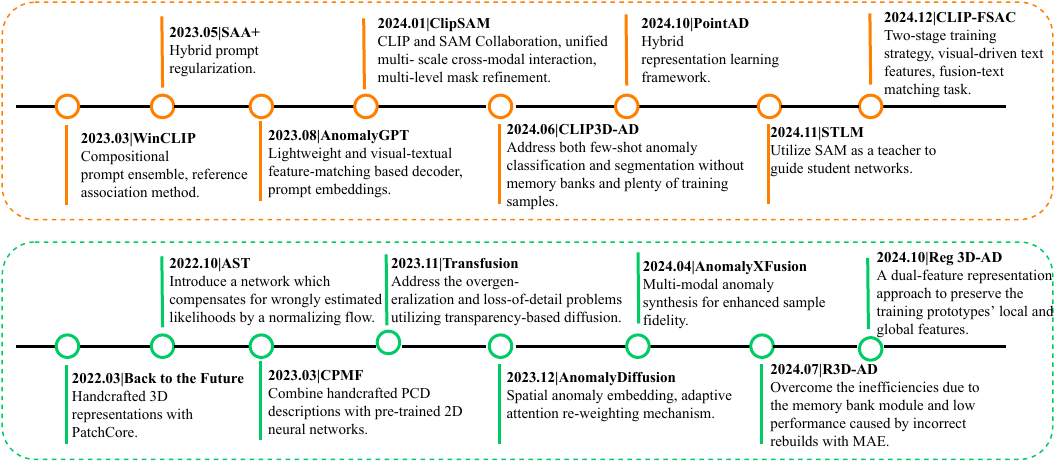}
    \caption{Representative methods along the development of FM and NFM models. The orange box illustrates the evolution of FM methods. WinCLIP introduced the use of prompt ensemble and multi-scale feature extraction with CLIP. Subsequently, SAA+ and Anomaly GPT incorporated SAM and GPT techniques, fostering the exploration of cross-modal approaches exemplified by ClipSAM. 3D FM methods emerged later, with CLIP3D-AD and PointAD focusing on addressing inconsistencies in multimodal data. Recently, 2D FM methods have achieved improvements in inference speed and accuracy, such as STLM based on a teacher-student framework and CLIP-FSAC employing vision-driven textual strategies.The green box presents the progression of NFM methods. The early method Back to the Future proposed handcrafted 3D representations but suffered from low efficiency and accuracy. Diffusion-based approaches, including TransFusion, AnomalyDiffusion, and AnomalyXFusion, effectively addressed these issues. In recent years, 3D generative techniques have been explored, with efforts concentrated on enhancing computational and storage efficiency.}
\end{figure*}

\subsection{2D SAM-Based Methods}
As a foundation model, the Segment Anything Model (SAM) \cite{kirillov2023sam,cao2023collaborative,wan2021industrial} has a powerful ability to extract high-quality segmentation masks. By leveraging large-scale pre-training data, it can perform instance segmentation on any object in various scenarios without the need for task-specific training. Consequently, SAM-based methods demonstrate good performance in zero-shot anomaly detection tasks. Cao et al. \cite{cao2023segment} utilize SAM and cascading prompt-guided object detection models \cite{liu2025grounding} to construct a vanilla baseline, i.e., Segment Any Anomaly (SAA). SAA generates preliminary anomaly regions through simple language prompts such as “defect” or “anomaly,” followed by a refinement process. They further introduce a mixed prompt regularization technique, enhancing the framework into Segment Any Anomaly+ (SAA+). To better process the masks generated by SAM, Li et al. propose ClipSAM \cite{li2024clipsam}, which combines the strengths of both CLIP and SAM. ClipSAM uses CLIP’s semantic understanding capabilities for anomaly localization and rough segmentation, then using the results as prompt constraints for SAM to refine the segmentation outcomes. In SAM-LAD \cite{peng2024sam}, Peng et al. introduce SAM to obtain object masks of the query and reference images. Each object mask is multiplied with the entire image’s feature map to obtain object feature maps, which are then used for object matching. Building on this, an Anomaly Measurement Model (AMM) is proposed to detect logical and structural anomalies. In UCAD \cite{liu2024unsupervised}, Liu et al. use SAM to enhance anomaly detection through Structure-based Contrastive Learning (SCL). By treating SAM-generated masks as structure, features within the same mask are drawn closer together, while others are pushed apart. This improves feature representation for anomaly detection. Li et al. \cite{li2024sam} propose a SAM-guided Two-stream Lightweight Model (STLM), prioritizing efficiency and mobile compatibility. One stream extracts features to distinguish normal from anomalous regions, while the other reconstructs anomaly-free images to enhance differentiation. With a shared mask decoder and feature aggregation, STLM delivers precise anomaly maps. In SPT \cite{yang2024spt}, Yang et al. introduced a Visual-Relation-Aware Adapter (VRA-Adapter) to help SAM better understand the relationships between different regions in the image, enhancing SAM's fine-grained understanding of anomaly patterns.

\subsection{2D CLIP-Based Methods}
CLIP-based methods \textbf{use large-scale pre-trained visual-language models that combine image coding with textual cues}, thereby strengthening the relationship between visual features and linguistic information, and perform particularly well in zero- and few-shot scenarios.

\subsubsection{Text Prompt}
Jeong et al. \cite{jeong2023winclip} propose a window-level anomaly detection method called WinCLIP, which achieves zero-shot anomaly segmentation \cite{radford2021learning} through combinatorial prompt ensemble and multi-scale feature extraction. Zhou et al. \cite{zhou2023anomalyclip} design an object-agnostic text prompt template to capture anomalous regions in an image by learning generic normality and abnormality prompts \cite{zhou2022conditional} and combining global and local contextual information. APRIL-GAN \cite{chen2023april} combines a text prompt integration strategy with a linear layer \cite{ross2017focal,milletari2016v} to improve the performance of zero and few-shot anomaly detection. SimCLIP reduces the reliance on hand-crafted prompts by introducing multi-level visual adapters with implicit prompt tuning. Qu et al. \cite{qu2024vcp} use visual contextual prompts to activate CLIP's anomaly semantic capability, eliminating the need for product-specific prompts. Cao et al. \cite{cao2025adaclip} propose to optimise zero-shot anomaly detection performance by combining static and dynamic learnable prompts. Li et al. \cite{li2024promptad} propose to convert normal prompts into anomaly prompts via semantic connectivity to build a large number of negative samples for prompt learning.

\subsubsection{Fine-Grained Alignment}
Gu et al. \cite{gu2024filo} improve the accuracy of anomaly localization by fine-grained local descriptions and optimised visual encoders. Zhang et al. \cite{zhang2024dual} propose to improve the accuracy of anomaly detection by using dual-image as visual references. Zuo et al. \cite{zuo2024clipfsac} propose that the performance of few-shot anomaly classification can be effectively improved by a two-stage training strategy and an image-to-text cross-attention module. Chen et al. \cite{chen2024clipad} propose to achieve fine-grained alignment through representative vector selection and a staged dual-path model. FiLo uses fine-grained description and high-quality localization to improve the accuracy and interpretability of anomaly detection. SAA+ \cite{cao2023segment} achieves more accurate anomaly localization through prompt-guided object detection and refinement techniques. ClipSAM \cite{li2024clipsam} improves zero-shot anomaly segmentation by combining the strengths of CLIP and SAM \cite{wang2024sam}, and leverages the semantic understanding capability of CLIP for anomaly localization and fine-grained segmentation \cite{wang2022cris}. Hu et al. propose a hierarchical freezing window self-attention mechanism \cite{xing2024less} that captures features at different levels by combining multi-level adapters for fine-grained localization \cite{sun2022dualcoop}. 

\subsection{2D GPT-Based Methods}
GPT-based methods \textbf{exploit the advantages of large-scale language models in natural language understanding and adaptive learning} to support the anomaly detection task by generating concrete textual descriptions, while being able to adapt their processing strategies to changing detection environments and anomaly types based on real-time input. 

AnomalyGPT \cite{gu2024anomalygpt} is the first to apply Large Vision-Language Models (LVLMs) \cite{ouyang2022training,touvron2023open} to anomaly detection tasks and supports multiple rounds of conversations, demonstrating excellent few-shot learning capabilities. Subsequently, Cao et al. \cite{cao2023towards} and xu et al. \cite{xu2024customizing} explore how to use LVLMs for general anomaly detection tasks across various domains. At the same time, they incorporate information from different modalities, such as domain knowledge, class context, and reference images as prompts to improve LVLMs' detection performance. Myriad \cite{li2023myriad} reduces the reliance on labelled data by combining visual experts with a large-scale multimodal model. Zhu et al. \cite{zhu2024alfa} propose a run-time prompt adaptation strategy to generate informative anomaly prompts, which combined with a fine-grained aligner can achieve accurate anomaly localization and enhance the dynamic adaptability of the model, making it more useful in diverse industrial scenarios. Zhang et al. \cite{zhang2024gpt} explore the potential of Visual Question Answering (VQA)-oriented GPT-4V (ision) in anomaly detecion \cite{cao2023towards}, introducing a GPT-4V-AD framework that integrates Granular Region Division, Prompt Designing, and Text2Segmentation. LogiCode \cite{zhang2024logicode} fully leverages the reasoning capabilities of LLMs. It extracts logical relationships from normal images and generates executable Python code to automatically detect logical anomalies in test images. The system also provides the specific location and detailed explanation of the anomalies. The innovative framework of LogiCode breaks through traditional anomaly detection methods, offering a more intelligent solution.

\subsection{3D CLIP-Based Methods}
In contrast to traditional 2D CLIP models that primarily process RGB images, \textbf{3D CLIP handles three-dimensional data—point clouds— which encompass more complex spatial structures and geometric information}. Currently, research in 3D anomaly detection is less developed compared to its 2D counterpart, largely because CLIP was initially trained on 2D RGB images paired with text. Consequently, 3D CLIP faces challenges in integrating point cloud data with images in a multimodal framework. Recent studies have made significant progress in overcoming the modality gap in 3D data processing, employing techniques such as multimodal noise reduction, multi-view processing and fusion of 3D data, as well as the integration of zero-shot learning to improve performance.

Zuo et al. \cite{zuo2024clip3d} proposed a multi-view fusion module that integrates 2D image features from different perspectives, thereby enhancing the representation capability of point cloud data and overcoming the challenges posed by modality differences when processing 3D point clouds directly. PointAD \cite{zhou2024pointad} achieves zero-shot 3D anomaly detection by rendering 3D point clouds from multiple views into 2D images \cite{zhang2022pointclip}, and then jointly optimizing 2D and 3D features through Hybrid Representation Learning. M3DM-NR \cite{wang2024m3dm} significantly improves data quality and reduces noise interference through a three-stage multimodal noise removal method. It leverages pre-trained CLIP and Point-BIND models, and employs multi-scale feature comparison and weighting to enhance the quality of training samples and improve the overall data purity.

\section{Non-Foundation Model Methods}
Unlike large-scale model-based methods that depend on extensive pretraining and complex multimodal fusion techniques, lightweight models improve detection accuracy through optimized architectures, feature extraction techniques, and computational efficiency. These methods are particularly suited for resource-limited scenarios that require fast inference, offering significant benefits for real-world deployment in industrial environments. This section will discuss the four main types of current lightweight model methods: statistical approaches, anomaly synthesis strategies, detection methods combining 2D RGB images with 3D point clouds, and 3D generation techniques. \textbf{The methods discussed in this section can be used as references for FM methods in the future, such as statistics-related methods, generation model, and data synthesis}.
Table 1 also shows the summary and overview of different methods based on NFM. Figure 4 presents the main timeline of NFM development.

\subsection{Statistics-Related Methods}
The statistical methods provide \textbf{an effective theoretical foundation} for improving the performance of anomaly detection models. Zhang et al. \cite{zhang2024sofs} propose a mixed normal Dice loss to improve the Dice loss. This loss function imposes a large penalty when the model predicts false positives, thus prioritizing the prevention of such incorrect predictions. Bae et al. \cite{bae2023pni} propose the PNI algorithm to address the impact of location and neighborhood information on the distribution of normal features. This algorithm employs a conditional probability based on neighborhood features, using a Multi-Layer Perceptron (MLP) network to model the distribution of normal features. Additionally, the method effectively captures positional information by constructing histograms of representative features at each location. LYU et al. \cite{lyu2024reb} consider variations in local feature density and propose the Local Density K-Nearest Neighbors (LDKNN) method to reduce the density bias in patch-level features. COAD \cite{qian2024friend} views overfitting as a controllable mechanism that enhances sensitivity to anomalies through controlled overfitting. It introduces the Aberrance Retention Quotient (ARQ) metric to precisely quantify the degree of overfitting, thereby identifying an optimal "golden overfitting interval" (the optimal ARQ) to optimize anomaly detection performance. BGAD \cite{yao2023bgad} designs a boundary-guided semi-push-pull (BG-SPP) loss. First, it generates an explicit boundary by learning the normal sample feature distribution. Based on this, it pulls together normal samples while pushing away anomalous samples, thereby strengthening the decision boundary. BGAD enables the model to effectively distinguish between seen and unseen anomalies using only a small number of anomalous samples. However, the scarcity of anomalous samples may still lead to inefficient feature learning, and BGAD does not fully address this key issue.

\subsection{Anomaly Synthesis Strategies}
Anomaly synthesis strategies aim to enhance the performance of anomaly detection models by \textbf{generating diverse and realistic abnormal samples}. Broadly, anomaly synthesis strategies can be categorized into the following types: 

\subsubsection{Generative Models}
Based on Denoising Diffusion Probabilistic Models (DDPM), Zhang et al. \cite{zhang2024realnet} introduce additional noise in the reverse diffusion process to control the intensity of the generated anomalous samples. Besides, Hu et al. \cite{hu2024anomalyxfusion} aggregate multiple modality features and integrate them into a unified embedding space, optimizing modality alignment. They then facilitate controlled generation through adaptive adjustments of the embedding based on diffusion steps. Jiang et al. \cite{jiang2024cagen} enhance the controllability of anomaly generation through fine-tuning a ControlNet model with text prompts and binary masks. Hu et al. \cite{hu2024anomalydiffusion} propose AnomalyDiffusion, which uses a Latent Diffusion Model (LDM) to generate anomalous images. It combines spatial anomaly embedding with an adaptive attention mechanism to improve the alignment between the generated anomalies and their corresponding masks. Li et al. \cite{li2024adabldm} build upon the Blended Latent Diffusion Model (BLDM) \cite{rombach2022high} with several innovations. They design a novel ’defect trimap’ to delineate the object masks and defect regions in generated images. They also introduce a cascaded ’editing’ stage in latent and pixel spaces to ensure structural coherence and detail fidelity. Additionally, they propose an online adaptation of the image encoder to further enhance image quality. Duan et al. \cite{duan2023dfmgan} train a data-efficient StyleGAN2 on defect-free images as the backbone. Then, they add defect-aware residual blocks to generate defect masks and manipulate the features within the masked regions, generating new defect images.

\subsubsection{Data Augmentation Techniques}
Based on normal features, Chen et al. \cite{chen2025glass} guide Gaussian noise through gradient ascent and truncated projection to synthesize weak anomalies around normal points. Besides, they create binary masks using Perlin noise and combine them with external textures to synthesize strong anomalies that are further away from normal points. Zhang et al. \cite{zhang2023destseg} generate anomalous images using Perlin noise and use them as input for the student network. By training the student network to remove the synthetic anomalous noise, they enhance the student network’s ability to represent features of anomalous samples, thereby improving the performance of the teacher-student framework in anomaly detection. Qin et al. \cite{qin2024cutswap} introduce semantic information for the generation of anomalous samples. They utilize LayerCAM to extract salient features from images and conduct clustering to identify the most significant regions. Subsequently, they select similar patch pairs and swap their positions. The negative samples generated in this way are more subtle yet realistic. Lin et al. \cite{lin2024split} develope a comprehensive anomaly simulation framework that combines reconstruction strategies for both transparent and opaque anomalies. By using selective augmentation and segmentation-based training strategies, they address the challenges of anomaly generation diversity, reconstruction quality, and overfitting. Bai et al. \cite{bai2024dfd} discover that small anomalies become more noticeable in the frequency domain. By transforming spatial images into multi-frequency representations, the discriminator learns joint representations between normal images and pseudo-anomalies, thereby improving the performance of few-shot anomaly detection. PBAS \cite{chen2024pbas} first learns a compact distribution of normal sample features with center constraints as an approximate decision boundary, which is used to guide the direction of feature-level anomaly synthesis. Then, it performs binary classification between the synthesized anomalies and normal features, further optimizing the decision boundary to ensure that the synthesized anomalies do not overlap with normal samples.

\subsection{Methods Combining 2D RGB and 3D Point Clouds}
The method of combining 2D RGB images with 3D point clouds improves the detection capabilities of traditional approaches, which are often limited by the lack of data from a single modality. \textbf{This is done by fusing features from both modalities: the rich color and texture features of 2D RGB images and the spatial and geometric information provided by 3D point clouds.}

Chu et al. propose a shape-guided expert-based learning framework that employs two expert models to detect anomalies in 3D structure and color appearance, respectively, and locates defects in test samples using a dual memory bank and shape-guided reasoning method. The model utilizes neural implicit functions (NIFs) \cite{ma2022surface} to represent local shapes and refines the complex structure of point clouds through signed distance fields, enabling point-level anomaly prediction. This significantly improves the accuracy of anomaly localization while reducing computational and memory costs. CPMF \cite{cao2024cpmf} generates pseudo-2D representations by projecting point clouds onto 2D and extracts semantic features using a pre-trained 2D neural network. These features complement 3D local features extracted from handcrafted point cloud descriptors and are unified into a global semantic and local geometric point cloud representation through feature alignment and fusion modules. Horwitz et al. \cite{horwitz2023back} highlighted that 3D methods are currently outperformed by 2D methods and proposed a solution combining rotation-invariant handcrafted feature representations with deep learning-based color features to improve 3D anomaly detection performance. TransFusion \cite{fuvcka2025transfusion} addresses the overgeneralization and detail loss issues by iteratively increasing the transparency of anomalous regions and gradually replacing them with the normal appearance while preserving the normal appearance of non-anomalous regions. Zavrtanik et al. \cite{zavrtanik20243dsr} introduced 3DSR, where DADA learns a universal discrete latent space that jointly models RGB and depth data. 3DSR performs discriminative anomaly detection in the feature space learned by DADA. M3DM constructs three separate memory banks for RGB, 3D, and fused features and performs anomaly detection by considering decisions from these memory banks through Decision Layer Fusion (DLF). To better align 3D point cloud features with 2D RGB features, Point Feature Alignment (PFA) was introduced. Rudolph et al. \cite{rudolph2023ast} presented the Asymmetric Student-Teacher Network (AST), which employs a normalized flow for density estimation as the teacher network and a conventional feed-forward network as the student network, solving the issue of insufficient output differences for anomalous data caused by similar student and teacher architectures in previous methods.

\subsection{3D Generation Methods}
\textbf{3D generative techniques use generative models to reconstruct normal samples or missing regions,} aiming to reduce computational overhead and improve model robustness, particularly addressing the challenges of model-free products and the difficulty in identifying missing regions.

Zhou et al. \cite{zhou2025r3d} employed a diffusion model-based data distribution transformation to completely mask abnormal geometries in the input, learning gradual displacement during the reverse diffusion process and explicitly controlling the reconstruction of abnormal shapes. Additionally, they proposed a 3D anomaly simulation strategy called Patch-Gen, designed to generate realistic defect shapes and bridge the gap between training and testing data. R3D-AD addresses challenges in 3D anomaly detection related to computational storage overhead and the detection of unmasked region anomalies. PointCore requires only a single memory bank to store local (coordinate) and global (PointMAE) representations, assigning different priorities to these local-global features to reduce computational costs and mitigate feature misalignment during inference. A ranking-based normalization method is used to eliminate distribution discrepancies between different anomaly scores, while the Iterative Closest Point (ICP) algorithm is applied to locally optimize point cloud registration results, enhancing decision robustness. Liu et al. \cite{liu2024uni} proposed a dual-branch structure where the feature-based branch and reconstruction-based branch detect surface defects and missing regions, respectively, with the latter incorporating Generative Adversarial Network Inversion (GAN-Inversion) for the first time to generate normal samples most similar to the input, thereby reducing false positives. Zhu et al. \cite{zhu2024group3ad} introduced the Inter-cluster Uniformity Network (IUN) and Intra-cluster Alignment Network (IAN), which respectively achieve inter-cluster dispersion and intra-cluster alignment in feature space, enhancing the uniformity and consistency of features. Moreover, the adaptive group center selection design focuses on regions with potential issues, prioritizing areas with significant local geometric changes, thereby improving the model's sensitivity.

\subsection{Conclusion and Outlooks}
This paper reviews the methodologies in industrial defect detection, focusing on FM approaches. Section 1 introduces the challenges posed by FM methods. In Section 2, we compare FM and NFM systematically. Section 3 reviews FM methods for 2D and 3D defect detection, while Section 4 summarizes NFM approaches.

Despite progress, several challenges remain, and further exploration is needed in the following areas:

\begin{itemize}
\item \textbf{Improving Detection Accuracy on Single-Scene Datasets:} While FM show impressive generalization across diverse scenarios, there is still a need to optimize their performance on specific scene datasets. Enhancing accuracy for a given dataset requires refining model fine-tuning processes, incorporating scene-specific features, and exploring specialized training techniques, such as transfer learning or domain adaptation. Further investigation into balancing model generalization and overfitting on limited datasets will be crucial to improving single-scene detection accuracy.

\item \textbf{Increasing Inference Speed in Few-Shot and Zero-Shot Scenarios:} FM, due to their extensive parameters, face challenges in inference speed, particularly in few-shot or zero-shot learning contexts. Speed improvement strategies, such as knowledge distillation, quantization, and model pruning, hold promise. Moreover, methods for optimizing inference, like efficient transfer of learned knowledge from large datasets to smaller ones or leveraging feature extraction techniques, could be explored to accelerate inference while maintaining accuracy.

\item \textbf{Enhancing 3D Detection Performance:} The performance of large models in 3D defect detection remains suboptimal, especially in single-scene scenarios. Improving 3D detection requires incorporating advanced 3D data processing methods, such as multi-view fusion, improved point cloud processing, and novel geometric feature extraction techniques. Additionally, coupling these methods with large models could enhance their ability to detect anomalies in complex 3D environments, where context and spatial relationships play a critical role.

\item \textbf{Synthetic Data for Specific 3D Scenarios:} Synthetic data generation, particularly for specific 3D industrial environments, could significantly boost FM performance in these scenarios. By generating diverse, realistic 3D defect samples through simulation or augmentation techniques, we can alleviate data scarcity and improve model robustness. Exploring the synergy between synthetic data and large models, especially in underrepresented or highly specialized 3D defect scenarios, could provide new avenues for training and fine-tuning defect detection models in real-world applications.
\end{itemize}

It is our hope that this survey provides a systematic summary and offers inspiration to readers for conducting research in related fields.

\bibliographystyle{IEEEtran}

\clearpage
\begin{appendices}
\appendix[RESOURCES]
We collect open-source information for FM and NFM methods, including the paper URL, code address (Github), and deep learning tools. Table 2 and Table 3 present the summarized information for FM and NFM methods respectively.

\begin{table*}
\centering
    \caption{A Collection of Published Papers and Codes for FM Methods.}
\resizebox{\textwidth}{!}{%
    \begin{tabular}{|c|c|c|c|} \hline 
         Methods&  Paper URL&  Code URL& Framework\\ \hline 
         ClipSAM \cite{li2024clipsam}&  \href{https://arxiv.org/pdf/2401.12665}{https://arxiv.org/pdf/2401.12665}&  \href{https://github.com/Lszcoding/ClipSAM}{https://github.com/Lszcoding/ClipSAM}& -\\ \hline 
         UCAD \cite{liu2024unsupervised}&  \href{https://arxiv.org/pdf/2401.01010}{https://arxiv.org/pdf/2401.01010}&  \href{https://github.com/shirowalker/UCAD}{https://github.com/shirowalker/UCAD}& PyTorch\\ \hline 
         SAM-LAD \cite{peng2024sam}& \href{https://arxiv.org/pdf/2406.00625}{https://arxiv.org/pdf/2406.00625}& -&-\\ \hline 
         SAA+ \cite{cao2023segment}& \href{https://arxiv.org/pdf/2305.10724}{https://arxiv.org/pdf/2305.10724}&  \href{https://github.com/caoyunkang/Segment-Any-Anomaly}{https://github.com/caoyunkang/Segment-Any-Anomaly}&PyTorch\\  \hline
         STLM \cite{li2024sam}& \href{https://arxiv.org/pdf/2402.19145}{https://arxiv.org/pdf/2402.19145}& \href{https://github.com/Qi5Lei/STLM}{https://github.com/Qi5Lei/STLM}&PyTorch\\\hline
          SPT \cite{yang2024spt}& \href{https://arxiv.org/pdf/2411.17217}{https://arxiv.org/pdf/2411.17217}& \href{https://github.com/THU-MIG/SAM-SPT}{https://github.com/THU-MIG/SAM-SPT}&-\\\hline
         WinCLIP \cite{jeong2023winclip}&\href{https://arxiv.org/pdf/2303.14814v1}{https://arxiv.org/pdf/2303.14814v1}&\href{https://github.com/openvinotoolkit/anomalib}{https://github.com/openvinotoolkit/anomalib}&PyTorch\\ \hline 
        AnoCLIP \cite{deng2023anovl}&\href{https://arxiv.org/pdf/2308.15939v2}{https://arxiv.org/pdf/2308.15939v2}&-&-\\ \hline 
        AnomalyCLIP \cite{zhou2023anomalyclip}&\href{https://arxiv.org/pdf/2310.18961v7}{https://arxiv.org/pdf/2310.18961v7}&\href{https://github.com/zqhang/anomalyclip}{https://github.com/zqhang/anomalyclip}&PyTorch\\ \hline
        AdaCLIP \cite{cao2025adaclip}&\href{https://arxiv.org/pdf/2407.15795v1}{https://arxiv.org/pdf/2407.15795v1}&\href{https://github.com/caoyunkang/adaclip}{https://github.com/caoyunkang/adaclip}&PyTorch\\ \hline 
        VCP-CLIP \cite{qu2024vcp}&\href{https://arxiv.org/pdf/2407.12276v1}{https://arxiv.org/pdf/2407.12276v1}&\href{https://github.com/xiaozhen228/vcp-clip}{https://github.com/xiaozhen228/vcp-clip}&PyTorch\\ \hline 
        SimCLIP \cite{deng2024simclip}&\href{https://openreview.net/pdf?id=kiH6PqRhwE}{https://openreview.net/pdf?id=kiH6PqRhwE}&\href{https://anonymous.4open.science/r/SimCLIP-CAEC}{https://anonymous.4open.science/r/SimCLIP-CAEC}&-\\ \hline 
        CLIP-AD \cite{chen2024clipad}&\href{https://arxiv.org/pdf/2311.00453v2}{https://arxiv.org/pdf/2311.00453v2}&-&-\\ \hline 
        CLIP-FSAC \cite{zuo2024clipfsac}&\href{https://www.ijcai.org/proceedings/2024/0203.pdf}{https://www.ijcai.org/proceedings/2024/0203.pdf}&-&-\\ \hline
        ClipSAM \cite{li2024clipsam}&\href{https://arxiv.org/pdf/2401.12665v2/2024/0203.pdf}{https://arxiv.org/pdf/2401.12665v2/2024/0203.pdf}&\href{https://github.com/lszcoding/clipsam}{https://github.com/lszcoding/clipsam}&-\\\hline
        SOWA \cite{hu2024sowa}&\href{https://arxiv.org/pdf/2407.03634v2}{https://arxiv.org/pdf/2407.03634v2}&\href{https://github.com/huzongxiang/sowa}{https://github.com/huzongxiang/sowa}&PyTorch\\\hline
        SAA+ \cite{cao2023segment}&\href{https://arxiv.org/pdf/2305.10724v1}{https://arxiv.org/pdf/2305.10724v1}&\href{https://github.com/caoyunkang/segment-any-anomaly}{https://github.com/caoyunkang/segment-any-anomaly}&PyTorch\\\hline
        APRIL-GAN \cite{chen2023april}&\href{https://arxiv.org/pdf/2305.17382v3}{https://arxiv.org/pdf/2305.17382v3}&\href{https://github.com/bychelsea/vand-april-gan}{https://github.com/bychelsea/vand-april-gan}&PyTorch\\\hline
        PromptAD \cite{li2024promptad}&\href{https://arxiv.org/pdf/2404.05231v2}{https://arxiv.org/pdf/2404.05231v2}&\href{https://github.com/funz-0/promptad}{https://github.com/funz-0/promptad}&PyTorch\\\hline
        FiLo \cite{gu2024filo}&\href{https://arxiv.org/pdf/2404.13671v2}{https://arxiv.org/pdf/2404.13671v2}&\href{https://github.com/casia-iva-lab/filo}{https://github.com/casia-iva-lab/filo}&PyTorch\\\hline
        Dual-Image Enhanced CLIP \cite{zhang2024dual}&\href{https://arxiv.org/pdf/2405.04782v1}{https://arxiv.org/pdf/2405.04782v1}&-&-\\\hline
        AnomalyGPT \cite{gu2024anomalygpt}&\href{https://arxiv.org/pdf/2308.15366v4}{https://arxiv.org/pdf/2308.15366v4}&\href{https://github.com/casia-iva-lab/anomalygpt}{https://github.com/casia-iva-lab/anomalygpt}&PyTorch\\\hline
        Myriad \cite{li2023myriad}&\href{https://arxiv.org/pdf/2310.19070v2}{https://arxiv.org/pdf/2310.19070v2}&-&-\\\hline
        ALFA \cite{zhu2024alfa}&\href{https://arxiv.org/pdf/2404.09654v2}{https://arxiv.org/pdf/2404.09654v2}&-&-\\ \hline
        GPT-4V-AD \cite{zhang2024gpt}& \href{https://arxiv.org/pdf/2311.02612}{https://arxiv.org/pdf/2311.02612}& \href{https://github.com/zhangzjn/GPT-4V-AD}{https://github.com/zhangzjn/GPT-4V-AD}&PyTorch\\\hline
        Customizable-VLM \cite{xu2024customizing}& \href{https://arxiv.org/pdf/2403.11083}{https://arxiv.org/pdf/2403.11083}& \href{https://github.com/Xiaohao-Xu/Customizable-VLM}{https://github.com/Xiaohao-Xu/Customizable-VLM}&PyTorch\\\hline
        LogiCode \cite{zhang2024logicode}& \href{https://arxiv.org/pdf/2406.04687}{https://arxiv.org/pdf/2406.04687}& -&-\\\hline
        CLIP3D-AD \cite{zuo2024clip3d}& \href{https://arxiv.org/pdf/2406.18941}{https://arxiv.org/pdf/2406.18941}& -&-\\ \hline
        PointAD \cite{zhou2024pointad}& \href{https://arxiv.org/pdf/2410.00320}{https://arxiv.org/pdf/2410.00320}& \href{https://github.com/zqhang/PointAD}{https://github.com/zqhang/PointAD}&PyTorch\\\hline
        M3DM-NR \cite{wang2024m3dm}& \href{https://arxiv.org/pdf/2406.02263}{https://arxiv.org/pdf/2406.02263}& -&-\\\hline
         Echo \cite{chen2025can}&\href{https://arxiv.org/pdf/2501.15795}{https://arxiv.org/pdf/2501.15795}& -&-\\\hline 
         KAnoCLIP \cite{li2025kanoclip}&\href{https://arxiv.org/pdf/2501.03786}{https://arxiv.org/pdf/2501.03786}&-&-\\\hline
    \end{tabular}
}
\end{table*}

\begin{table*}
\centering
    \caption{A Collection of Published Papers and Codes for NFM Methods.}
\resizebox{\textwidth}{!}{%
\begin{tabular}{|c|c|c|c|} \hline 
    Methods&  Paper URL&  Code URL& Framework\\ \hline 
    SOFS \cite{zhang2024sofs}& \href{https://arxiv.org/pdf/2407.21351}{https://arxiv.org/pdf/2407.21351}& \href{https://github.com/zhangzilongc/SOFS}{https://github.com/zhangzilongc/SOFS}&PyTorch\\\hline
     PNI \cite{bae2023pni}& \href{https://arxiv.org/pdf/2211.12634}{https://arxiv.org/pdf/2211.12634}& \href{https://github.com/wogur110/PNI_Anomaly_Detection}{https://github.com/wogur110/PNI\_Anomaly\_Detection}&PyTorch\\\hline
     REB \cite{lyu2024reb}& \href{https://arxiv.org/pdf/2308.12577}{https://arxiv.org/pdf/2308.12577}& \href{https://github.com/ShuaiLYU/REB}{https://github.com/ShuaiLYU/REB}&PyTorch\\\hline
     BGAD \cite{yao2023bgad}& \href{https://arxiv.org/pdf/2207.01463}{https://arxiv.org/pdf/2207.01463}& \href{https://github.com/xcyao00/BGAD}{https://github.com/xcyao00/BGAD}&PyTorch\\\hline
     COAD \cite{qian2024friend}& \href{https://arxiv.org/pdf/2412.06510}{https://arxiv.org/pdf/2412.06510}& -&-\\\hline
     GLASS \cite{chen2025glass}& \href{https://arxiv.org/pdf/2407.09359}{https://arxiv.org/pdf/2407.09359}& \href{https://github.com/cqylunlun/GLASS}{https://github.com/cqylunlun/GLASS}&PyTorch\\\hline
     AdaBLDM \cite{li2024adabldm}& \href{https://arxiv.org/pdf/2402.19330}{https://arxiv.org/pdf/2402.19330}& \href{https://github.com/GrandpaXun242/AdaBLDM.git}{https://github.com/GrandpaXun242/AdaBLDM.git}&PyTorch\\\hline
     RealNet \cite{zhang2024realnet}& \href{https://arxiv.org/pdf/2403.05897}{https://arxiv.org/pdf/2403.05897}& \href{https://github.com/cnulab/RealNet}{https://github.com/cnulab/RealNet}&PyTorch\\\hline
     CAGEN \cite{jiang2024cagen}& \href{https://ieeexplore.ieee.org/document/10447663}{https://ieeexplore.ieee.org/document/10447663}& -&-\\\hline
     AnomalyXFusion \cite{hu2024anomalyxfusion}& \href{https://arxiv.org/pdf/2404.19444}{https://arxiv.org/pdf/2404.19444}& \href{https://github.com/hujiecpp/MVTec-Caption}{https://github.com/hujiecpp/MVTec-Caption}&-\\\hline
     AnomalyDiffusion \cite{hu2024anomalydiffusion}& \href{https://arxiv.org/pdf/2312.05767}{https://arxiv.org/pdf/2312.05767}& \href{https://github.com/sjtuplayer/anomalydiffusion}{https://github.com/sjtuplayer/anomalydiffusion}&PyTorch\\\hline
     DFMGAN \cite{duan2023dfmgan}& \href{https://arxiv.org/pdf/2303.02389}{https://arxiv.org/pdf/2303.02389}& \href{https://github.com/Ldhlwh/DFMGAN}{https://github.com/Ldhlwh/DFMGAN}&PyTorch\\\hline
     DeSTSeg \cite{zhang2023destseg}& \href{https://arxiv.org/pdf/2211.11317}{https://arxiv.org/pdf/2211.11317}& -&-\\\hline
     CutSwap \cite{qin2024cutswap}& \href{https://arxiv.org/pdf/2311.18332}{https://arxiv.org/pdf/2311.18332}& -&-\\\hline
     Split Training \cite{lin2024split}& \href{https://arxiv.org/pdf/2308.15068}{https://arxiv.org/pdf/2308.15068}& -&-\\\hline
     DFD \cite{bai2024dfd}& \href{https://arxiv.org/pdf/2403.04151}{https://arxiv.org/pdf/2403.04151}& \href{https://github.com/yuhbai/DFD}{https://github.com/yuhbai/DFD}&PyTorch\\\hline
     PBAS \cite{chen2024pbas}& \href{https://ieeexplore.ieee.org/stamp/stamp.jsp?tp=\&arnumber=10716437}{https://ieeexplore.ieee.org/stamp/stamp.jsp?tp=\&arnumber=10716437}& \href{https://github.com/cqylunlun/PBAS}{https://github.com/cqylunlun/PBAS}&PyTorch\\\hline
    Shape-Guided \cite{chu2023shape}&  \href{https://openreview.net/pdf?id=IkSGn9fcPz}{https://openreview.net/pdf?id=IkSGn9fcPz}&  \href{https://github.com/jayliu0313/Shape-Guided}{https://github.com/jayliu0313/Shape-Guided}& PyTorch\\ \hline 
    CPMF \cite{cao2024cpmf}&  \href{https://arxiv.org/pdf/2303.13194v1}{https://arxiv.org/pdf/2303.13194v1}&  \href{https://github.com/caoyunkang/CPMF}{https://github.com/caoyunkang/CPMF}& PyTorch\\ \hline 
    Back to the Feature \cite{horwitz2023back}&  \href{https://arxiv.org/pdf/2203.05550}{https://arxiv.org/pdf/2203.05550}&  \href{https://github.com/eliahuhorwitz/3D-ADS}{https://github.com/eliahuhorwitz/3D-ADS}& PyTorch\\ \hline 
    TransFusion \cite{fuvcka2025transfusion}&  \href{https://arxiv.org/pdf/2311.09999v2}{https://arxiv.org/pdf/2311.09999v2}&  \href{https://github.com/maticfuc/eccv_transfusion}{https://github.com/maticfuc/eccv\_transfusion}& PyTorch\\ \hline 
    3DSR \cite{zavrtanik20243dsr}&  \href{https://arxiv.org/pdf/2311.01117v1}{https://arxiv.org/pdf/2311.01117v1}&  \href{https://github.com/vitjanz/3dsr}{https://github.com/vitjanz/3dsr}& PyTorch\\ \hline 
    M3DM \cite{wang2024m3dm}& \href{https://arxiv.org/pdf/2303.00601v2}{https://arxiv.org/pdf/2303.00601v2}& \href{https://github.com/nomewang/m3dm}{https://github.com/nomewang/m3dm}&PyTorch\\ \hline 
    AST \cite{rudolph2023ast}& \href{https://arxiv.org/pdf/2210.07829v2}{https://arxiv.org/pdf/2210.07829v2}& \href{https://github.com/marco-rudolph/ast}{https://github.com/marco-rudolph/ast}&PyTorch\\ \hline 
    R3D-AD \cite{zhou2025r3d}& \href{https://arxiv.org/pdf/2407.10862v1}{https://arxiv.org/pdf/2407.10862v1}&  -&-\\ \hline
    Reg 3D-AD \cite{liu2024real3d}& \href{https://arxiv.org/pdf/2309.13226}{https://arxiv.org/pdf/2309.13226}& \href{https://github.com/M-3LAB/Real3D-AD}{https://github.com/M-3LAB/Real3D-AD}&PyTorch\\\hline
    PointCore \cite{zhao2024pointcore}& \href{https://arxiv.org/pdf/2403.01804v1}{https://arxiv.org/pdf/2403.01804v1}& -&-\\ \hline 
    Uni-3DAD \cite{liu2024uni}& \href{https://arxiv.org/pdf/2408.16201}{https://arxiv.org/pdf/2408.16201}& -&-\\ \hline
    Group3AD \cite{zhu2024group3ad}& \href{https://arxiv.org/pdf/2408.04604}{https://arxiv.org/pdf/2408.04604}& -&-\\ \hline
    PVQAE \cite{cheng2024patch}&\href{https://arxiv.org/pdf/2501.09187}{https://arxiv.org/pdf/2501.09187}&-&-\\\hline 
    CAI \cite{wang2025stones}&\href{https://arxiv.org/pdf/2501.15211}{https://arxiv.org/pdf/2501.15211}&-&-\\\hline 
 \end{tabular}
 }
\end{table*}

\end{appendices}
\end{document}